\documentclass[11pt]{article}
\usepackage[preprint]{acl}
\usepackage{amsmath}
\usepackage{booktabs}
\usepackage{enumitem}
\usepackage{tcolorbox}
\tcbuselibrary{listings, breakable}
\usepackage{enumitem}
\usepackage{times}
\usepackage{latexsym}
\usepackage{amssymb}
\usepackage{subcaption}
\usepackage[T1]{fontenc}
\usepackage[utf8]{inputenc}

\usepackage{microtype}

\usepackage{inconsolata}

\usepackage{graphicx}

\title{MAESTRO: Meta-learning Adaptive Estimation of Scalarization Trade-offs for Reward Optimization}

\author{Yang Zhao$^{\spadesuit}\footnotemark[1]$,Hepeng Wang$^{\spadesuit}\footnotemark[1]$, Xiao Ding$^{\spadesuit}\footnotemark[2]$,Yangou Ouyang$^{\spadesuit}$,Bibo Cai$^{\spadesuit}$, Kai Xiong$^{\spadesuit}$,\\
Jinglong Gao$^{\spadesuit}$, Zhouhao Sun$^{\spadesuit}$, Li Du$^{\heartsuit}$, Bing Qin$^{\spadesuit}$ and  Ting Liu$^{\spadesuit}$ \\
  $^{\spadesuit}$Research Center for Social Computing and Interactive Robotics,\\ %
  Harbin Institute of Technology, China \\
  $^{\heartsuit}$Beijing Academy of Artificial Intelligence, Beijing, China\\
 \texttt{\{yangzhao,hpwang\}@ir.hit.edu.cn}\\\
 \\\\}

\begin{document}

\maketitle
\renewcommand*{\thefootnote}{\fnsymbol{footnote}}
\footnotetext[1]{These authors contributed equally to this work.}
\renewcommand*{\thefootnote}
{\fnsymbol{footnote}}
\footnotetext[2]{Corresponding Author.}
\renewcommand*{\thefootnote}
{\fnsymbol{footnote}}
\renewcommand*{\thefootnote}
{\arabic{footnote}}
\maketitle
\begin{abstract} Group-Relative Policy Optimization (GRPO) has emerged as an efficient paradigm for aligning Large Language Models (LLMs), yet its efficacy is primarily confined to domains with verifiable ground truths. Extending GRPO to \textbf{open-domain settings} remains a critical challenge, as \textbf{unconstrained generation} entails multi-faceted and often conflicting objectives—such as creativity versus factuality—where rigid, static reward scalarization is inherently suboptimal. To address this, we propose \textbf{MAESTRO} (\textbf{M}eta-learning \textbf{A}daptive \textbf{E}stimation of \textbf{S}calarization \textbf{T}rade-offs for \textbf{R}eward \textbf{O}ptimization), which introduces a meta-cognitive orchestration layer that treats reward scalarization as a dynamic latent policy, leveraging the model's terminal hidden states as a semantic bottleneck to perceive task-specific priorities. We formulate this as a contextual bandit problem within a bi-level optimization framework, where a lightweight Conductor network co-evolves with the policy by utilizing group-relative advantages as a meta-reward signal. Across seven benchmarks, MAESTRO consistently outperforms single-reward and static multi-objective baselines, while preserving the efficiency advantages of GRPO, and in some settings even reducing redundant generation. The source code for this work is publicly available at: https://github.com/zy125413/MAESTRO.

\end{abstract}

\section{Introduction}
\begin{figure}[t]
    \centering
    \includegraphics[width=0.9\linewidth]{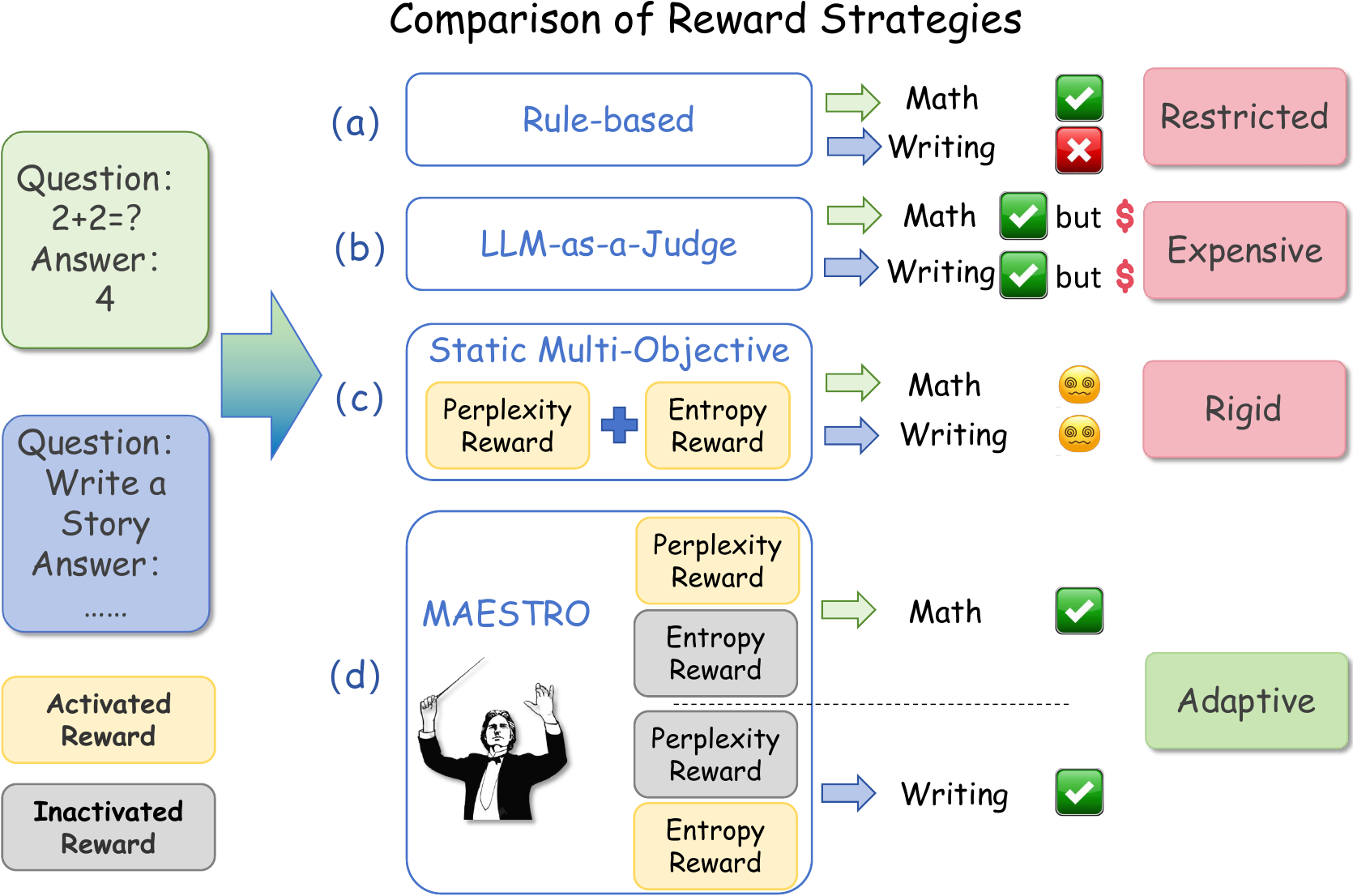}
    \caption{Comparison of reward orchestration strategies. 
    (a) Rule-based methods, limited to verifiable ground truths. 
    (b) LLM-as-a-Judge, flexible but with high computational overhead. 
    (c) Static multi-objective optimization using fixed, prompt-independent reward weights across all contexts. 
    (d) MAESTRO (Ours), which dynamically adapts reward weights according to prompt--response semantics.}
    \label{fig:overview0}
\end{figure}

GRPO~\citep{shao2024deepseekmath} has been widely adopted as a powerful paradigm for aligning LLMs, particularly in complex reasoning tasks~\cite{guo2025deepseek,openr1}. By computing relative advantages against the empirical mean of group-wise rewards, GRPO dramatically enhances performance in domains with verifiable ground truths, such as mathematics and code generation~\citep{guo2025deepseek,yue2025vapoefficientreliablereinforcement}. However, extending this success to open-ended generation remains a significant challenge~\cite{liu2025nover}. The absence of objective verification rules creates a ``reward vacuum'', where standard rule-based scalarization fails to capture the nuanced, context-dependent objectives of open-ended language generation (Figure~\ref{fig:overview0}(a)).

To bridge this gap, current open-domain alignment primarily bypasses verifiable rules through two distinct paradigms: (1) Model-based evaluators utilize auxiliary LLMs as judges~\citep{zheng2023judging} or verifiers (Figure~\ref{fig:overview0}(b))~\citep{liu2025skywork}. While versatile, they incur prohibitive computational overhead from repeated large-model inference~\citep{zheng2023judging}, and introduce inherent stylistic biases—such as favoring longer responses or specific sentence structures—that skew alignment~\citep{wang2023aligning}. (2) Intrinsic heuristic rewards leverage statistical proxies like reasoning perplexity~\citep{liu2025nover} or token entropy~\citep{agarwal2025unreasonable}. However, these proxies often correlate poorly with nuanced human utility; for instance, a generic platitude may yield lower perplexity than an informative response. Furthermore, these heuristics rely on static, context-agnostic scalarization, obscuring fine-grained trade-offs between heterogeneous desiderata such as factuality and creativity.
This failure to capture generative nuances underscores that open-domain alignment is \textbf{structurally a Multi-Objective Optimization (MOO) problem} \citep{zhao2025pareto}. Existing strategies, by relying on \textbf{static scalarization} \citep{taechoyotin2025remor}, effectively collapse this high-dimensional frontier into a single, fixed coordinate. As contrastively illustrated in Figure~\ref{fig:overview0}(c) and (d), such rigidity is inherently suboptimal: while mathematical reasoning demands strict logical rigor, creative writing requires a strategic shift toward stylistic diversity. 

In this paper, we propose \textbf{MAESTRO}, a framework that addresses open-domain alignment by reformulating reward orchestration as a \textbf{contextual bandit} problem~\citep{li2010contextual}. 
Rather than relying on static scalarization, MAESTRO treats reward weighting as a context-dependent decision, enabling adaptive trade-offs across heterogeneous objectives.
To instantiate this formulation, we leverage the observation that terminal hidden states serve as a \textbf{semantic bottleneck}~\citep{voita2019bottom, andreas-etal-2018-learning}, encoding high-level information about task intent and generation characteristics. 
We use these latent representations as the \emph{context} for a lightweight meta-policy, termed the \emph{Conductor}, which selects a reward scalarization strategy for each prompt--response pair.
MAESTRO is optimized through a \textbf{bi-level optimization} scheme. 
In the inner loop, the policy model ($\pi_\theta$) is trained with GRPO under the selected reward configuration, while in the outer loop, the Conductor ($\pi_\phi$) is updated using GRPO group-relative advantage as a meta-reward signal, reinforcing reward configurations that lead to stronger policy improvements. A key challenge in this setting is \emph{meta-credit assignment}: under group-relative normalization, naively coupling reward orchestration with policy updates can lead to unstable or degenerate meta-gradients. 
MAESTRO addresses this challenge via asynchronous, two-time-scale updates that decouple the optimization of the Conductor from token-level policy training, substantially improving stability. 
Together, these components enable the co-evolution of adaptive reward landscapes and emergent policy behaviors, allowing MAESTRO to dynamically adapt evaluative criteria to the context-dependent demands of open-ended generation.

We evaluate MAESTRO using mainstream LLMs (e.g., Qwen3-8B~\cite{qwen3technicalreport} and Llama-3.1-8B-Instruct~\cite{dubey2024llama}) across seven diverse open-domain datasets. Empirical results show that MAESTRO consistently outperforms state-of-the-art methods like Nover~\citep{liu2025nover}. Furthermore, our analysis shows that dynamic reward orchestration preserves the efficiency of GRPO, and in some cases we even observe up to a 20.1\% improvement in training throughput due to reduced redundant generation.

Our contributions are summarized as follows: 
\begin{itemize}[nosep,leftmargin=*] 
\item \textbf{Contextual Reward Orchestration Framework:} 
We propose \textbf{MAESTRO}, a meta-learning framework that formulates reward orchestration as a contextual bandit problem within GRPO, using terminal hidden states to adapt reward trade-offs per trajectory.

\item \textbf{Advantage-Driven Bi-level Optimization:} 
We formulate reward scalarization as a bi-level optimization problem, and leverage GRPO group-relative advantages as a meta-learning signal for reward orchestration, addressing meta-credit assignment in group-relative settings.

\item \textbf{Efficiency-Preserving Open-Domain Alignment:} 
Across seven benchmarks, MAESTRO achieves superior alignment performance while preserving—and in some cases improving—the efficiency advantages of GRPO.

\end{itemize}

\section{Related Work}

\paragraph{Reinforcement Learning for LLMs} 
Traditional PPO~\cite{schulman2017proximal, ouyang2022training} incurs substantial memory overhead due to centralized critics and reference models. Rule-based RL, such as GRPO~\citep{shao2024deepseekmath,guo2025deepseek} and its variants~\citep{liu2025understandingr1zeroliketrainingcritical,yu2025dapo}, sidesteps these requirements via group-wise relative comparisons. However, these methods typically rely on monolithic rewards unsuitable for nuanced open-domain tasks. MAESTRO adapts this efficient paradigm to open-ended generation by replacing static rules with a learned meta-controller for non-deterministic alignment~\citep{liu2025nover}.

\begin{figure*}[t]
    \centering
    \includegraphics[width=0.85\linewidth]{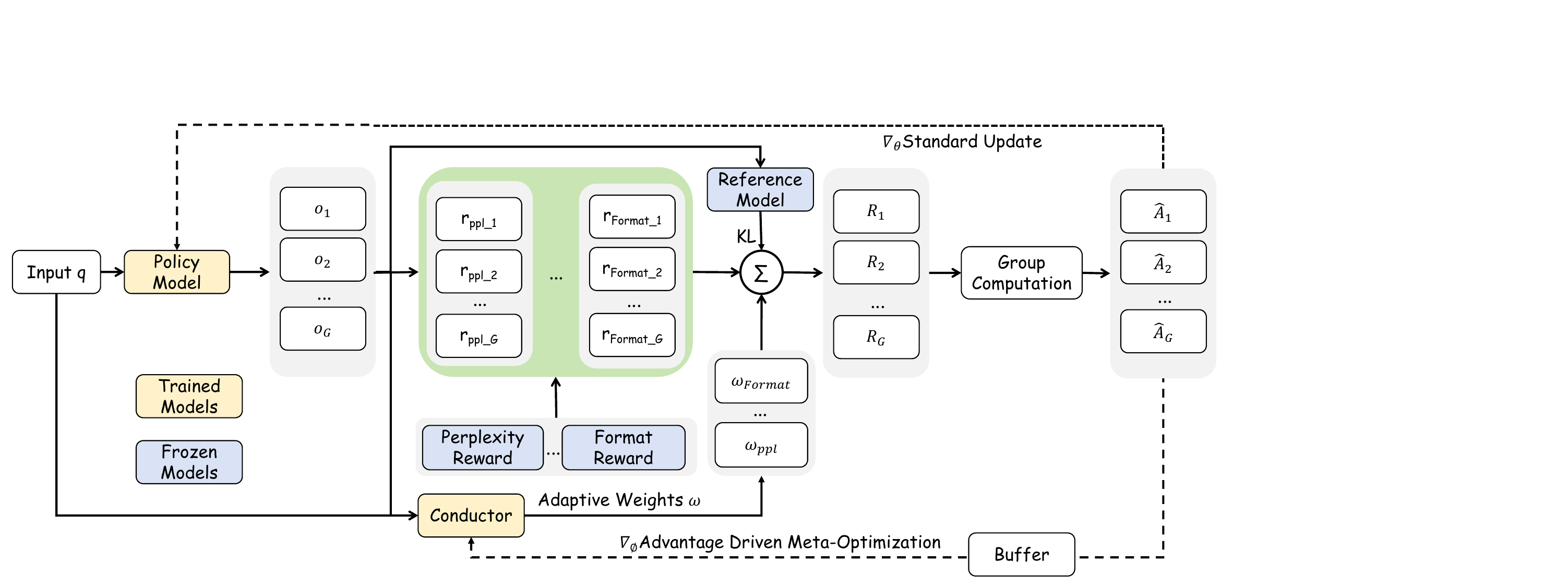}
    \caption{Architecture of the MAESTRO framework. Given a prompt $q$, the policy model $\pi_\theta$ samples a group of candidate outputs $\{o_i\}$. The Conductor $\pi_\phi$ simultaneously processes prompt--response hidden states to sample a reward-emphasis action $a$, inducing a weighting scheme $\mathbf{w}^{(a)}$. Raw reward vectors $\mathbf{r}$ and KL penalties are fused by the scalarization node ($\Sigma$) to compute the scalar reward $R$, which is normalized via group computation to obtain group-relative advantages $\hat{A}$. MAESTRO adopts bi-level optimization: $\pi_\theta$ is updated with GRPO, while $\pi_\phi$ performs meta-updates using group advantages as reinforcement signals.}
    \label{fig:overview}
\end{figure*}

\paragraph{Open-Domain Alignment with Heuristic Rewards}
Lacking verifiable signals, recent works employ heuristic proxies like perplexity~\citep{liu2025nover}, entropy~\citep{agarwal2025unreasonable}, or LLM-as-a-Judge~\citep{zheng2023judging}. Yet, single-metric heuristics impose rigid biases prone to entropy collapse~\citep{yu2025dapo}, while model-based judges introduce prohibitive latency. Unlike these singular or expensive evaluators, MAESTRO dynamically orchestrates multiple low-cost signals to balance semantic quality with computational throughput.

\paragraph{Multi-Objective Optimization for LLMs}
Multi-objective reinforcement learning (MORL) for LLM alignment has been explored from multiple perspectives.
Single-policy approaches aggregate objectives via static scalarization~\citep{taechoyotin2025remor, li2025optimizing}, which are simple but inflexible, while multi-policy or Pareto-based methods approximate the Pareto frontier using evolutionary or gradient-level optimization~\citep{callaghan2025extending,li2025gradient}, at the cost of training and maintaining multiple large-scale policies~\citep{mei2025ai}.
Motivated by these limitations, we formulate reward orchestration as a meta-policy approach with a bi-level structure, a direction identified as promising yet underexplored for MORL in LLMs~\citep{mei2025ai}.
Related bandit-based methods such as DYNAOPT~\citep{min-etal-2024-dynamic} adjust reward weights at a global training-stage level based on aggregate performance statistics, which is primarily suited for domain-specific settings with relatively homogeneous objectives, whereas MAESTRO performs instance-level, semantic-conditioned reward orchestration for open-domain LLM alignment and is optimized via advantage-driven bi-level learning tailored to group-relative optimization.

\section{Methodology: MAESTRO}

We propose \textbf{MAESTRO} (\textbf{M}eta-learning \textbf{A}daptive \textbf{E}stimation of \textbf{S}calarization \textbf{T}rade-offs for \textbf{R}eward \textbf{O}ptimization), a framework for open-domain alignment that dynamically orchestrates trade-offs among heterogeneous reward objectives via meta-learning.
MAESTRO is formulated as a bi-level optimization framework and consists of two tightly coupled components.
First, we introduce a lightweight \emph{Conductor} that observes semantic representations of prompt--response trajectories and selects context-dependent reward scalarizations, cast as a \textbf{contextual bandit} problem (\S\ref{sec:conductor}). 
Second, we propose an advantage-driven meta-optimization scheme that leverages GRPO group-relative advantages to stably train the Conductor alongside the policy model (\S\ref{sec:Advantage-Driven Meta-Optimization}).

\subsection{Conductor Network}
\label{sec:conductor}

\paragraph{Contextual Semantic Representation}
We define the context $h \in \mathbb{R}^{d_{\text{model}}}$ as the last hidden state of policy $\pi_\theta$ after processing the full sequence (prompt and response). This design is grounded in the \textbf{progressive abstraction} property of Transformers, where the final layer serves as a \textbf{semantic bottleneck} crystallizing the comprehension of task intent and linguistic realization~\citep{voita2019bottom, andreas-etal-2018-learning}. At this stage, high-level semantic features---such as logical rigor, complexity, and structural constraints---are explicitly decoded, providing sufficient statistics for instance-specific reward modulation. We use $h_{i,j}$ to denote the joint prompt--response representation $h(x_i, y_{i,j})$ throughout the remainder of this section. 

\paragraph{The Conductor Architecture}
Leveraging the linear separability of terminal semantics, the Conductor $f_\phi$ is implemented as a lightweight linear projection head. Given a representation $h$, the Conductor produces a categorical distribution over reward-emphasis actions:
\begin{equation}
\pi_\phi(\cdot \mid h) = \mathrm{softmax}\left((W_\phi h + b_\phi) / \tau\right).
\end{equation}

To facilitate stable meta-learning, the Conductor employs \textbf{sampling from a categorical mode distribution} during training. Each action $a \sim \pi_\phi$ induces a specific reward-emphasis mode, providing discrete gradients for effective credit assignment. This sampling-based exploration allows the Conductor to probe diverse regions of the Pareto frontier, which matures into a smooth, continuous preference structure during inference.

\subsection{Advantage-Driven Meta-Optimization}
\label{sec:Advantage-Driven Meta-Optimization}

We optimize the Conductor using the policy improvement dynamics of the backbone model. Specifically, we define the meta-objective $J(\phi)$ as the expected GRPO advantage under rollout-conditioned reward emphases:
\begin{equation}
J(\phi) = \mathbb{E}_{(x, y, a) \sim P_{\phi,\theta}} \left[ \hat{A}(x, y; w(h,a)) \right],
\end{equation}
where \(w(h,a) \in \Delta^{K-1}\) denotes the continuous reward-scalarization vector induced by the sampled discrete action \(a\), serving as a meta-level abstraction for stable credit assignment, with \(x \sim \mathcal{D}\), \(y \sim \pi_\theta(\cdot \mid x)\), and \(a \sim \pi_\phi(\cdot \mid h(x,y))\). The discrete training objective is described in detail in Appendix~\ref{sec:onehot}.

\paragraph{Meta-Credit Assignment under Group-Relative Advantages}
A naive approach to reward orchestration assigns a fixed, prompt-level weight vector to all rollouts within a GRPO group. However, due to the \textit{baseline-invariant} nature of group-relative advantages, such uniform weighting can lead to signal degeneracy: since advantages are mean-centered within each group, the resulting meta-gradient may vanish, leaving the Conductor with little ability to distinguish effective reward configurations.
In practice, MAESTRO adopts \textbf{intra-group heterogeneous sampling} as a simple mechanism to expose informative variance under group-relative advantages. For a given prompt $x_i$, the Conductor samples an independent reward-emphasis action $a_{i,j} \sim \pi_\phi$ for each response $y_{i,j}$ in the group, inducing a form of \textbf{meta-competition} across rollouts. This breaks the symmetry of the group baseline and provides the variance necessary for learning context-dependent reward trade-offs.

\paragraph{Asynchronous Meta-Updates}
MAESTRO employs a \textbf{two-time-scale asynchronous update scheme} to decouple the meta-optimization from the primary policy training loop. We buffer the $(h_{i,j}, a_{i,j}, \hat{A}_{i,j})$ triplets collected during GRPO training and periodically update the Conductor $\phi$ via the Policy Gradient Theorem:
\begin{equation}
\begin{aligned}
\hat{\nabla}_\phi J(\phi) = \frac{1}{NG} \sum_{i,j} \Big[ & \hat{A}_{i,j} \nabla_\phi \log \pi_\phi(a_{i,j} \mid h_{i,j}) \\
& + \lambda_{\text{ent}} \nabla_\phi \mathcal{H}(\pi_\phi) \Big],
\end{aligned}
\end{equation}
where $N$ and $G$ denote the meta-batch and group sizes, respectively; $\hat{A}_{i,j}$ is the group-relative advantage; $a_{i,j}$ represents the sampled reward-emphasis action; $h_{i,j}$ is the contextual semantic representation; and $\lambda_{\text{ent}}$ scales the entropy regularization $\mathcal{H}$ to encourage exploration of the preference space. At inference time, the Conductor bypasses discrete sampling and directly outputs the continuous distribution $\pi_\phi(\cdot \mid h)$, providing a deterministic and smooth integration of reward signals that encapsulates the learned preference structure.

\section{Experiments and Analysis}

\subsection{Experimental Setup}
\label{sec:exp_setup}

\paragraph{Tasks and Datasets}

Following prior work on open-domain RL~\citep{liu2025nover}, we evaluate MAESTRO across seven open-domain benchmarks spanning four task
categories: (i) \emph{General Reasoning} (Natural Reasoning~\citep{yuan2025naturalreasoning}, GeneralThoughts~\citep{huggingface-dataset-GT},
WebInstruct~\citep{huggingface-dataset-WI}), (ii) \emph{Creative Writing} (SS-GEN~\citep{feng2025ss}), (iii) \emph{Social
Intelligence} (EmoBench~\citep{sabour2024emobench}, ToMBench~\citep{chen2024tombench}), and (iv) \emph{Multilingual Generation}
(OPUS-Books~\citep{tiedemann2012parallel}). All datasets are split into
train--test sets following standard protocols used in prior work.
Additional preprocessing details are provided in Appendix~\ref{sec:appendix_data}.

\paragraph{Reward Specification}

Following prior work on open-domain reinforcement learning~\citep{liu2025nover}, we instantiate MAESTRO with a heterogeneous reward space $\mathcal{R} \subset \mathbb{R}^K$ comprising $K=5$ components. 
Detailed mathematical formulations are provided in Appendix~\ref{sec:appendix_rewards}.
These reward signals are designed to capture different aspects of generation quality.
Specifically, we employ a \textbf{perplexity-based reward} ($r_{\mathrm{ppl}}$) as a proxy for reasoning consistency, together with a \textbf{format validity reward} ($r_{\mathrm{fmt}}$) to enforce structural constraints. 
To balance exploration and verbosity, we incorporate an \textbf{entropy reward} ($r_{\mathrm{ent}}$) and a \textbf{length penalty} ($r_{\mathrm{len}}$). 
Finally, we include a \textbf{semantic preference reward} ($r_{\mathrm{pref}}$) derived from a pretrained reward model such as Skywork-Reward~\citep{liu2025skywork} to capture high-level qualities including helpfulness and coherence.

\paragraph{Training and Evaluating}

We run experiments on two backbone LLMs: Qwen3-8B~\citep{qwen3technicalreport} and
Llama-3.1-8B-Instruct~\citep{dubey2024llama}. All methods are trained using full-parameter
fine-tuning under the GRPO framework implemented in TRL~\citep{vonwerra2022trl}.
To ensure fair comparison, all GRPO-based baselines share identical
training hyperparameters within each task.
Details of hyperparameters and Conductor update schedules are deferred
to Appendix~\ref{sec:appendix_training}. To ensure an unbiased evaluation and mitigate self-rewarding bias, we adopt a cross-model evaluation protocol. 
For reasoning-oriented datasets with relatively well-defined target
answers (Natural Reasoning, GeneralThoughts, WebInstruct, EmoBench, and
ToMBench), we use Qwen3-235B-A32B as the evaluation judge. For SS-GEN and OPUS-Books, we use Gemini-2.5-Flash~\citep{comanici2025gemini25pushingfrontier} due to its stronger performance in
long-form narrative generation and multilingual evaluation.

\paragraph{Baselines}
We compare MAESTRO against a comprehensive set of baselines:
\textbf{Base} (pretrained model), \textbf{CoT} (chain-of-thought prompting),
\textbf{SFT}, and several GRPO variants.
\textbf{NOVER}~\citep{liu2025nover} applies GRPO with conditional perplexity as the primary reward,
while \textbf{EM-GRPO} follows entropy-minimization approaches in prior work~\citep{agarwal2025unreasonable} and applies a negative entropy reward within the GRPO framework.
To isolate the effect of dynamic reward composition, we additionally include
two ablations: \textbf{Equal-Weights}, which uses a fixed uniform reward
scalarization, and \textbf{Random-Weights}, which samples a random weight
vector per response. Our full method predicts prompt-dependent reward
weights using the Conductor.

\begin{table*}[t]
\centering
\small
\setlength{\tabcolsep}{4.5pt}
\begin{tabular}{lcccccccc}
\toprule
& \multicolumn{8}{c}{\textbf{Qwen3-8B}} \\
\cmidrule(lr){2-9}
\textbf{Dataset} 
& \textbf{Base} (\%) 
& \textbf{SFT} (\%) 
& \textbf{CoT} (\%) 
& \textbf{NOVER} (\%) 
& \textbf{Eq} (\%) 
& \textbf{EM} (\%) 
& \textbf{Rand} (\%) 
& \textbf{Ours} (\%)  \\
\midrule
NR              
& 39.60 & 26.00 & 43.00 & 46.90  & 47.20 & \underline{52.00} & 45.50& \textbf{53.20} \\
SS-GEN          
& 33.10 & 68.70  & 76.77 & 77.76& \underline{90.55} & 88.78 & 90.35 & \textbf{92.52}  \\
WebInstruct     
& 7.80  & 34.60 & 25.70 & 42.70 & 34.90 & \underline{43.40} & 40.10 & \textbf{43.50} \\
ToMBench        
& 5.72  & 46.92 & 25.79 & 56.24  & 38.27 & 63.83 & \underline{65.34} & \textbf{71.88}\\
GeneralThoughts 
& 34.00 & 34.70 & 34.20 & 64.60  & 66.90 & \underline{68.00} & 35.70 & \textbf{68.10}\\
OPUS-Books      
& 5.10  & 5.50  & 6.60  & 10.10  & 11.00 & \underline{11.70} & 11.40 & \textbf{12.60} \\
EmoBench        
& 36.71 & \underline{46.09} & 44.53 & 42.19 & 42.97 & 41.40 & 42.19 &\textbf{47.66} \\
\midrule
& \multicolumn{8}{c}{\textbf{Llama-3.1-8B-Instruct}} \\
\cmidrule(lr){2-9}
\textbf{Dataset} 
& \textbf{Base} (\%) 
& \textbf{SFT} (\%) 
& \textbf{CoT} (\%) 
& \textbf{NOVER} (\%) 
& \textbf{Eq} (\%) 
& \textbf{EM} (\%) 
& \textbf{Rand} (\%) 
& \textbf{Ours} (\%)  \\
\midrule
NR              
& 9.90   & 13.30   & 18.70 & 19.00    & 9.90 & \underline{19.80} & 19.20 & \textbf{21.30}\\
SS-GEN          
& 11.81   & 33.46   & 20.07 & 36.81    & \underline{58.86} & 54.92 & 51.18 & \textbf{60.43}\\
WebInstruct     
& 3.10   & 18.40   & 28.90 & 29.20   & 28.20 & \underline{30.20} & 28.50  & \textbf{30.30}\\
ToMBench        
& 3.27   & 64.88   & 63.12 & 64.29   & 65.34 & 65.23 & \underline{65.93} & \textbf{66.86} \\
GeneralThoughts 
& 11.30   & 19.90   & 29.00 & 29.10   & \underline{32.50} & 30.10 & 27.20 & \textbf{34.90} \\
OPUS-Books      
& 3.80   & 3.90   & 8.10  & 7.10   & 7.30 & \underline{10.20} & 8.10 & \textbf{11.80} \\
EmoBench        
& 3.13   & 34.38   & 42.97 & 42.97    & \underline{43.75} & \underline{43.75} & 42.19& \textbf{44.53} \\
\bottomrule
\end{tabular}
\caption{Main results across all datasets. Scores represent the preference percentage (\%) evaluated by an external LLM judge via a 3-way majority vote. All models are fine-tuned on the same backbone using identical GRPO hyperparameters. To ensure a fair comparison, \textbf{Eq}, \textbf{Rand}, and \textbf{Ours} all utilize the same set of five reward functions, differing only in their orchestration: Eq employs fixed uniform weights; Rand assigns random weights per response; and Ours utilizes the proposed Conductor for prompt-dependent dynamic weighting. \textbf{EM} represents the entropy-minimization baseline. Bold and underline denote the best and (possibly tied) second-best results, respectively.
}
\label{tab:main-results}
\end{table*}

\subsection{Main Results}

Table~\ref{tab:main-results} summarizes performance across all benchmarks.
MAESTRO achieves the strongest overall results on both backbone models,
consistently outperforming pretrained models, supervised fine-tuning,
chain-of-thought prompting, and GRPO-based baselines.
These gains are observed across diverse task categories, including reasoning,
creative writing, social intelligence, and multilingual generation, suggesting
that MAESTRO generalizes well beyond any single domain.
Notably, MAESTRO consistently improves upon static or heuristic reward
compositions (e.g., NOVER, EM-GRPO, and fixed-weight ablations), highlighting the
benefit of dynamically adapting reward trade-offs to different prompts.

\paragraph{Multi-objective Rewards Outperform Single-reward}
Single-reward GRPO variants exhibit strong but task-dependent behavior:
NOVER performs well on structured reasoning benchmarks, while EM-GRPO can
be competitive on some reasoning tasks.
However, both approaches degrade on open-ended domains and fail to
generalize across task categories.
By integrating multiple complementary reward signals, MAESTRO consistently
improves over strong single-reward baselines, indicating that open-domain
alignment benefits from balancing multiple objectives rather than
optimizing a single inductive bias.

\paragraph{Dynamic Reward Weighting is Essential}
Fixed uniform weighting (\textbf{Eq}) and randomly sampled weighting
(\textbf{Rand}) yield moderate gains over single-reward baselines but
remain inferior to MAESTRO.
Despite using the same reward components, MAESTRO surpasses Eq and Rand
on nearly all tasks (e.g., substantial gains on Natural Reasoning, ToMBench, and SS-GEN), demonstrating that optimal reward trade-offs are
instance-dependent and must be learned rather than fixed.

\paragraph{Robustness Across Models and Domains}
MAESTRO’s improvements hold consistently across both Qwen3-8B and
Llama-3.1-8B-Instruct backbones, spanning reasoning, creative writing,
social intelligence, and multilingual generation.
This robustness suggests that MAESTRO captures general reward trade-off
adaptation principles rather than exploiting dataset-specific heuristics.

\subsection{Analysis}

In this section, we analyze (i) why single-objective GRPO variants succeed only on subsets of tasks, (ii) what reward trade-offs the Conductor learns across domains, and (iii) whether MAESTRO introduces practical overhead.

\paragraph{Why EM-GRPO Excels on Reasoning Tasks}

Interestingly, EM-GRPO, which optimizes a negative entropy objective, performs strongly on several reasoning-centric benchmarks. As shown in Table~\ref{tab:main-results}, it closely matches MAESTRO on datasets such as \textsc{NR}, \textsc{WebInstruct}, and \textsc{GeneralThoughts}. This is consistent with prior findings~\citep{agarwal2025unreasonable} that low-entropy decoding encourages deterministic and concise reasoning while suppressing irrelevant hypotheses.
While reasoning tasks benefit from an entropy-minimization bias for deterministic logic, this ``one-size-fits-all'' approach induces mode collapse in open-ended generation. MAESTRO circumvents this by treating inductive biases as pluggable, context-dependent rewards rather than global constraints. However, this bias becomes misaligned in open-ended or socially grounded tasks (e.g., \textsc{ToMBench} and \textsc{SS-GEN}), which demand richer expression and flexible structure. As a result, EM-GRPO degrades substantially on such domains, illustrating that a single monotonic objective cannot generalize across heterogeneous tasks. These observations motivate dynamic reward composition. Rather than enforcing a global inductive bias, MAESTRO adaptively emphasizes entropy-sensitive objectives only when appropriate, capturing task-dependent reward trade-offs automatically.

\paragraph{Training-time Overhead}
We evaluate whether MAESTRO preserves the efficiency of GRPO on three representative datasets spanning different generation regimes: SS-GEN (long-form generation), ToMBench (short responses) and WebInstruct (medium-length structured generation). Given that GRPO training time is primarily dominated by model rollouts, these datasets collectively cover short-, medium-, and long-horizon generation settings.
Table~\ref{tab:training-time} reports the total training time with and without the Conductor. As illustrated, MAESTRO introduces a minimal computational footprint; on WebInstruct, the overhead is a marginal +4.0\%. Conversely, on SS-GEN, MAESTRO achieves a significant 20.1\% speedup. This efficiency gain stems from the Conductor's ability to learn dynamic reward trade-offs that discourage redundant or excessively verbose outputs early in the rollout phase, effectively reducing the average sequence length. Overall, MAESTRO maintains the efficiency advantages of GRPO, remaining highly competitive in structured tasks and even providing substantial speedups in long-form generation settings.

\begin{table}[h]
\centering
\small
\setlength{\tabcolsep}{4pt}
\begin{tabular}{lccc}
\toprule
\textbf{Dataset} & \textbf{w/ Cond.} & \textbf{w/o Cond.} & \textbf{$\Delta$ Time} \\
\midrule
SS-GEN        & \texttt{05:20:04} & \texttt{06:40:25} & \textbf{-20.1}\% \\
ToMBench      & \texttt{01:18:00} & \texttt{01:12:47} & +6.7\% \\
WebInstruct   & \texttt{02:40:05} & \texttt{02:33:52} & +4.0\% \\
\bottomrule
\end{tabular}
\caption{
Training time comparison with and without the Conductor on Qwen3-8B.
$\Delta$ Time is computed relative to the setting without the Conductor.
}
\label{tab:training-time}
\end{table}

\paragraph{Comparison with LLM-as-a-Judge Reward}
LLM-as-a-Judge has emerged as a prominent paradigm for reward modeling, relying on large language models to evaluate and rank candidate outputs based on a manually specified evaluation prompt. 
However, prior work has shown that the resulting reward signal is highly sensitive to prompt design: overly strict prompts may overlook valid aspects of a response, while overly permissive ones can introduce noisy or irrelevant criteria~\citep{liu2025nover}.
To evaluate this paradigm, we implement a GRPO baseline where a Qwen3-32B model serves as a judge to rank rollout candidates.
As shown in Table~\ref{tab:lj-comparison}, MAESTRO consistently outperforms this baseline across all evaluated tasks. This difference stems from a key methodological distinction. LLM-as-a-Judge relies on a fixed, prompt-defined evaluation policy, whereas MAESTRO learns context-dependent reward weightings over multiple specialized signals, yielding a more stable and informative optimization signal.
\begin{table}[h]
\centering
\small
\setlength{\tabcolsep}{4pt}
\begin{tabular}{lccc}
\toprule
\textbf{Dataset} & \textbf{MAESTRO} & \textbf{LLM-as-a-Judge} \\
\midrule
NR           & \textbf{53.20} &  52.50 \\
SS-GEN       & \textbf{92.52} &  91.93 \\
WebInstruct  & \textbf{43.50} &  43.30 \\
\bottomrule
\end{tabular}
\caption{
Comparison between MAESTRO and an LLM-as-a-Judge reward baseline. The baseline utilizes a smaller Qwen3-32B as a Reward Baseline Judge to provide scalar signals during GRPO training.
}
\label{tab:lj-comparison}
\end{table}

\paragraph{Reward Weight Patterns Across Task Categories}

\begin{figure}[t]
    \centering
\begin{subfigure}[t]{\linewidth}
    \centering
    \includegraphics[width=0.9\linewidth,trim=0 4.5cm 0 0,clip]{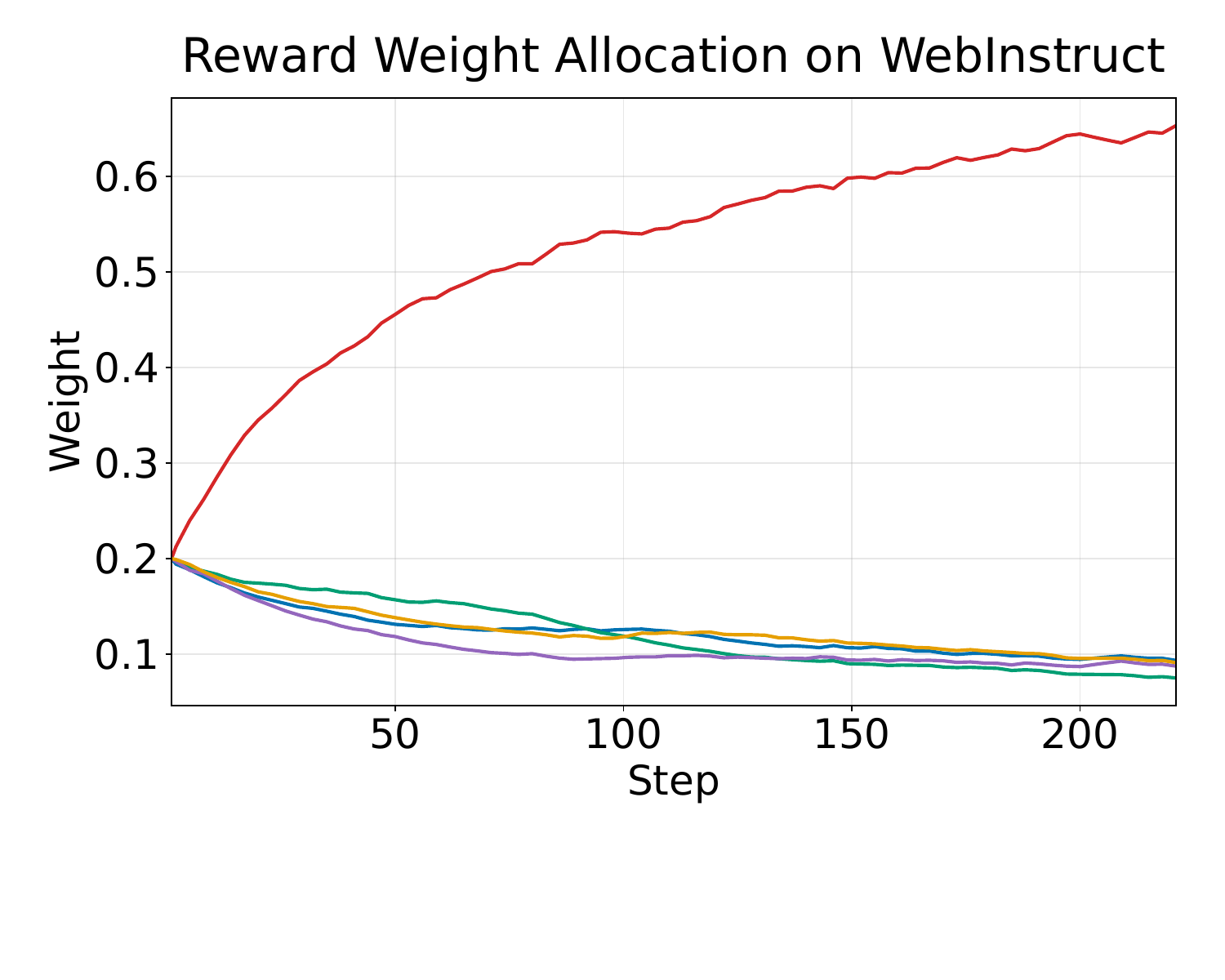}
    \label{fig:reward-ssgen}
\end{subfigure}
\vspace{-0.8em}
\begin{subfigure}[t]{\linewidth}
    \centering
    \includegraphics[width=0.9\linewidth]{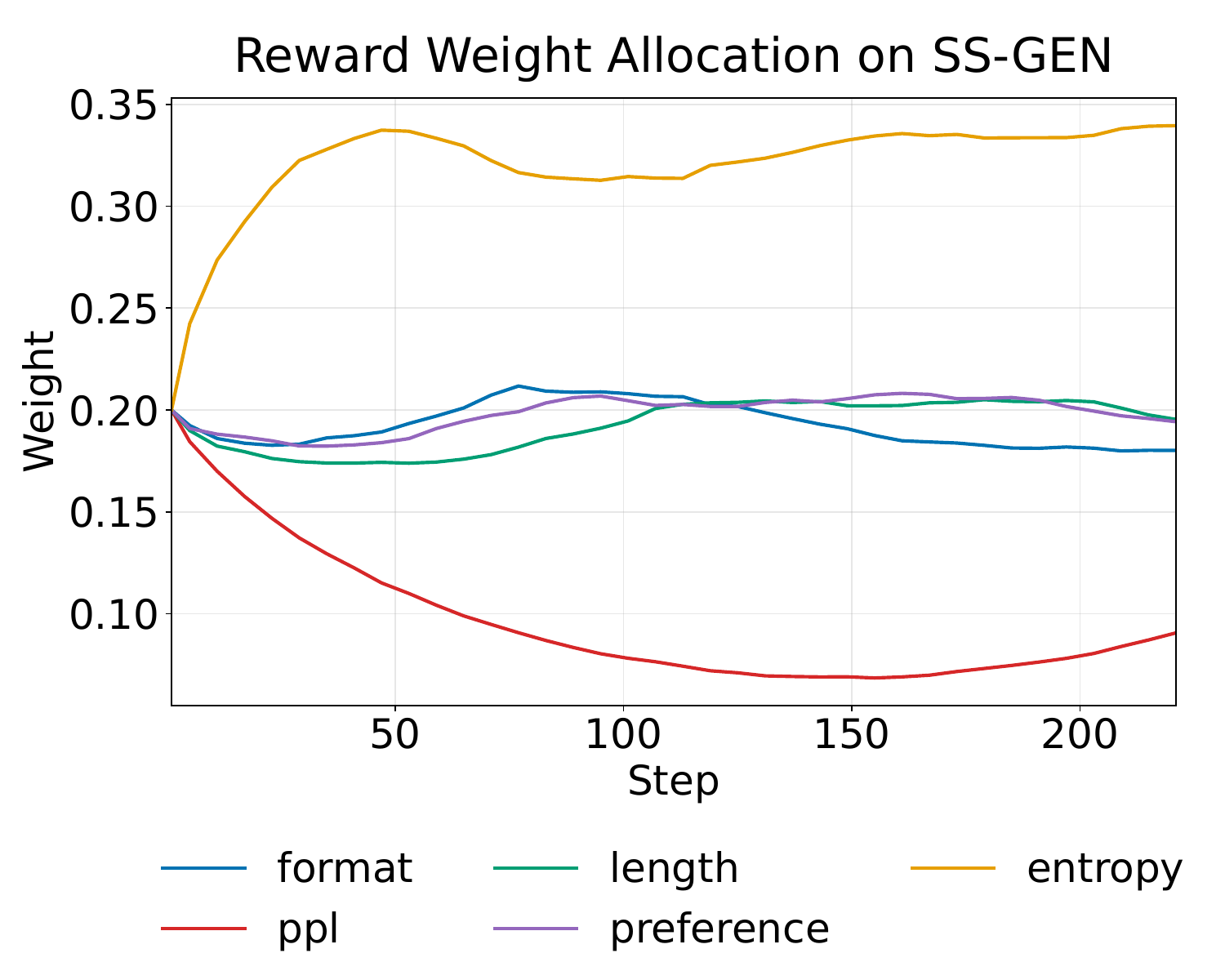}
    \label{fig:reward-webinstruct}
\end{subfigure}

    \caption{
    Comparison of learned reward weight allocation dynamics in the course of training across two datasets.
    Both plots use the same linear training-step axis to facilitate direct comparison.
    }
    \label{fig:reward-weight-dynamics}
\end{figure}

To better understand how MAESTRO adapts reward trade-offs, we analyze the reward weight distributions predicted by the Conductor on representative tasks from different categories. Figure~\ref{fig:reward-weight-dynamics} visualizes the evolution of reward weights during training on two datasets with very different styles: SS-GEN (creative writing) and WebInstruct (structured reasoning), respectively.
On SS-GEN, the Conductor assigns the highest weight to the entropy reward, encouraging exploration and expressive diversity, while substantially down-weighting the reasoning (perplexity-based) reward. Preference-model signals retain a moderate but non-dominant weight, providing weak semantic guidance without enforcing a single correct answer. This allocation reflects the open-ended nature of creative writing and illustrates why objectives that favor low-entropy or strict correctness, such as EM-GRPO, generalize poorly to such domains. WebInstruct exhibits a simple and decisive reward allocation pattern.
As a structured multiple-choice reasoning task, the Conductor assigns a dominant weight to the  reasoning reward ($r_{\text{ppl}}$) while substantially down-weighting all other signals. This reflects the correctness-driven nature of the task, where auxiliary objectives contribute little useful feedback.
Notably, these task-specific patterns emerge rapidly during training and then stabilize, providing behavioral evidence that MAESTRO learns meaningful and robust reward trade-offs aligned with task semantics rather than applying a fixed or arbitrary weighting scheme.

\section{Ablation Studies}
\label{sec:ablation}

We conduct a set of ablation experiments to evaluate the design choices
behind our Conductor and to understand which components are essential
for stable and effective dynamic reward composition.

\paragraph{Effect of Conductor Input Representation}
MAESTRO assumes reward composition depends on the model's internal comprehension of semantics rather than surface statistics. To verify this, we vary the layer for extracting $h(x,y)$. As shown in Table~\ref{tab:ablation-position}, last-layer embeddings (\textsc{Last}) perform best, while earlier representations show noticeable degradation. This gap confirms that dynamic orchestration relies on task understanding crystallized in upper layers. Our findings align with \citet{tenney2019you}, where high-level conceptual features localize in terminal representations. Lower layers lack the ``self-perception'' to distinguish task priorities like reasoning or creativity, justifying MAESTRO's operation over abstract latent states.

\begin{table}[h]
\centering
\small
\setlength{\tabcolsep}{4pt}
\begin{tabular}{lccc}
\toprule
\textbf{Dataset} & \textbf{Last} (\%) & \textbf{First} (\%) & \textbf{Middle} (\%) \\
\midrule
SS-GEN            & \textbf{92.52} & 91.30 & 88.80 \\
ToMBench          & \textbf{71.88} & 69.90 & 64.41 \\
GeneralThoughts   & \textbf{68.10} & 64.20 & 63.20 \\
OPUS-Books        & \textbf{12.60} & \textbf{12.60} & 11.10 \\
\bottomrule
\end{tabular}
\caption{
Ablation on Conductor input representation (Qwen3-8B).
\textsc{Last}/\textsc{Middle}/\textsc{First} use the prompt+response embedding from the Last / middle / first LLM layer as Conductor input.}
\label{tab:ablation-position}
\end{table}

\paragraph{Effect of Meta-loss Formulation} MAESTRO utilizes the GRPO advantage as a meta-reward to reinforce effective reward configurations. To test the robustness of this advantage-driven formulation, we ablate entropy regularization and evaluate pairwise perturbations to the REINFORCE update. While both variants incur moderate performance drops (Table~\ref{tab:ablation-loss}), they consistently outperform static baselines with fixed or random weights. This confirms that advantage-weighted gradients are the primary learning signals for the Conductor, while entropy provides stability. These results validate the use of GRPO advantages for meta-learning and demonstrate that MAESTRO's reward composition is robust rather than sensitive to loss configurations.

\begin{table}[h]
\centering
\small
\setlength{\tabcolsep}{4pt}
\begin{tabular}{lccc}
\toprule
\textbf{Dataset} & \textbf{Ours} (\%) & \textbf{-Entropy} (\%) & \textbf{+Pairwise} (\%) \\
\midrule
NR           & \textbf{53.20} & 48.40 &  49.50 \\
SS-GEN       & \textbf{92.52} & 91.70 &  91.73 \\
WebInstruct  & \textbf{43.50} & 42.50 &  42.00 \\
\bottomrule
\end{tabular}
\caption{ Ablation on Conductor training loss (Qwen3-8B). \textsc{Ours}: full meta-loss; \textsc{-Entropy}: removes entropy regularization; \textsc{+Pairwise}: adds pairwise perturbation to the REINFORCE objective. }
\label{tab:ablation-loss}
\end{table}

\paragraph{Asynchronous Meta-updates Provide Stable Improvements} MAESTRO employs a two-time-scale optimization scheme, where the LLM is updated at the token level while the Conductor learns prompt-level preferences.
This separation allows the Conductor to optimize against aggregated rollout-level advantages, rather than noisy token-level gradients that fluctuate across individual updates.
We compare this asynchronous scheme against joint training, where the Conductor is updated via direct backpropagation through GRPO.
As shown in Table~\ref{tab:joint}, joint training degrades performance, whereas asynchronous updates yield consistent gains.
We attribute this to a mismatch in optimization time scales: coupling the Conductor to token-level gradients entangles long-horizon preferences with short-term stochastic updates, obscuring reliable credit assignment.
In contrast, buffering prompt-level advantages smooths transient noise and enables the Conductor to learn stable, context-dependent reward trade-offs over time.

\begin{table}[h]
\centering
\small
\setlength{\tabcolsep}{4pt}
\begin{tabular}{lccc}
\toprule
\textbf{Dataset} & \textbf{Async-Conductor} (\%) & \textbf{Joint-Conductor} (\%) \\
\midrule
NR        & \textbf{53.20} &  50.00 \\
ToMBench  & \textbf{71.88} &  68.50 \\
SS-GEN    & \textbf{92.52} &  80.71 \\
\bottomrule
\end{tabular}
\caption{
Comparison of asynchronous and joint Conductor training strategies
(Qwen3-8B).
\textsc{Async-Conductor} updates the Conductor asynchronously using buffered
rollout-level advantages, while \textsc{Joint-Conductor} updates it jointly
with the LLM by backpropagating through GRPO.
}
\label{tab:joint}
\end{table}

\section{Conclusion}
In this paper, we introduced \textbf{MAESTRO}, a meta-learning framework that dynamically adapts reward scalarization for open-domain LLM alignment. By formulating reward weighting as a contextual bandit problem, MAESTRO leverages the optimization advantage from GRPO to learn a ``Conductor'' policy, \textbf{conditioned on terminal hidden states,} that modulates the trade-off between conflicting objectives such as rigor, creativity, and format validity. Extensive experiments on seven diverse benchmarks demonstrate that MAESTRO significantly outperforms static weighting schemes and single-reward baselines, \textbf{while maintaining training efficiency}. Our findings underscore that open-ended generation is inherently a multi-objective problem where the optimal reward composition is highly context-dependent. MAESTRO provides a computationally efficient and robust solution to this challenge, extending the success of rule-based RL to broader, non-deterministic domains.

\section{Limitations}
While MAESTRO achieves strong performance, several limitations remain. First, although the Conductor adds negligible training overhead, the framework still relies on the quality of the underlying reward components; if the constituent reward signals (e.g., the semantic preference model) are flawed or biased, dynamic weighting can only partially mitigate these issues. Second, our current exploration is limited to scalarization of five specific reward types; exploring a broader or automatically discovered reward space remains future work. Finally, while we validate our method on 8B-parameter models due to compute constraints, scaling the Conductor to larger dense models or mixture-of-experts (MoE) architectures requires further empirical verification.

\section{Acknowledgements}
The research in this article is supported by the New Generation Artificial Intelligence of China (2024YFE0203700), National Natural Science Foundation of China under Grants U22B2059 and 62576124.

\bibliography{custom}

\appendix

\section{Theoretical Analysis: Conductor vs. Critic and Counter-Hacking Mechanisms}
\label{sec:appendix_theoretical}

To ensure a rigorous understanding of the MAESTRO framework, we provide a detailed analysis of the Conductor's role, its fundamental differences from the PPO Critic, and the structural constraints that prevent reward hacking.

\subsection{Delineation from the PPO Critic}
A common misconception is to view the Conductor as a traditional value function (Critic) used in PPO. We explicitly delineate these two components across three dimensions:

\begin{itemize}[leftmargin=*]
    \item \textbf{Output Space and Semantics:} The PPO Critic estimates a scalar value $V(s) \in \mathbb{R}$, predicting the expected return. In contrast, the Conductor is a meta-policy $\pi_\phi(a|h)$ that outputs a probability distribution over the \textbf{reward weight simplex} $\Delta^{K-1}$. It defines \textit{how} the model should be evaluated for a specific context, rather than \textit{what} the score will be.
    \item \textbf{Learning Objective:} The Critic is trained via supervised regression to minimize the Mean Squared Error (MSE) of value predictions. The Conductor is trained via an advantage-driven meta-objective to maximize the resolution of the policy improvement signal ($\hat{A}$). Its goal is not to ``fit'' the rewards but to ``steer'' the optimization trajectory toward the most productive Pareto trade-offs.
    \item \textbf{Role in Optimization:} While a Critic collapses all reward dimensions into a single scalar for variance reduction, the Conductor preserves the multi-dimensional nature of the reward space. It treats reward scalarization as a dynamic latent decision, acting as a \textbf{strategic orchestrator} rather than a passive value accountant.
\end{itemize}

\subsection{Robustness to Reward Hacking}
A central concern in meta-learning reward orchestration is whether the Conductor might trivially assign high weights to reward components where the policy already achieves high scores (self-rewarding bias). MAESTRO is structurally robust to such hacking due to the following mechanisms:

\paragraph{1. The Zero-Sum Constraint of GRPO Advantages}
The Conductor is optimized using the GRPO advantage $\hat{A}_{i,j}$, which is mean-centered by definition:
\begin{equation}
\hat{A}_{i,j} = \frac{R_{i,j} - \text{mean}(R_i)}{\text{std}(R_i)}
\end{equation}
Since $\mathbb{E}[\hat{A}] \approx 0$ for any group, the Conductor cannot increase its meta-reward by simply inflating the scalarized reward $R$. If all samples in a group perform equally well on a ``easy'' metric, the resulting advantage will be zero. To obtain positive meta-gradients, the Conductor is forced to identify reward configurations that \textbf{maximize the differentiation} (signal resolution) between superior and inferior trajectories within the same context.

\paragraph{2. Signal-to-Noise Ratio (SNR) Optimization}
The Conductor acts as a contextual filter. If it assigns high weight to a ``noisy'' reward component (e.g., entropy in a deterministic logic task), the resulting $\hat{A}_{i,j}$ will correlate poorly with the policy's actual improvement, leading to a meta-penalty. It is thus incentivized to prioritize components that provide the cleanest improvement signal for the given task.

\paragraph{3. Meta-Credit Assignment via Heterogeneous Sampling}
By sampling independent orchestration actions $a_{i,j}$ for each rollout within a group, MAESTRO induces a \textbf{meta-competition} between different scalarization schemes. An orchestration policy is reinforced only if its specific weight configuration leads to higher relative advantages for ``better'' samples. This ensures the Conductor's decisions are grounded in actual policy performance gains.

\section{Reward Definitions}
\label{sec:appendix_rewards}

MAESTRO employs a heterogeneous reward space consisting of five
complementary components. Each reward captures a distinct aspect of
generation quality. All reward values are normalized to the range
$[0,1]$ before scalarization.

\paragraph{Format Validity Reward ($r_{\mathrm{fmt}}$)}
We enforce a predefined response structure that separates intermediate
reasoning and final answers into distinct segments.
The format reward is defined as:
\[
r_{\mathrm{fmt}}(y) =
\begin{cases}
1, & \text{if } y \text{ satisfies the required format}, \\
0, & \text{otherwise}.
\end{cases}
\]

\paragraph{Perplexity Reward ($r_{\mathrm{ppl}}$)}
The perplexity reward measures the logical consistency of the model's
reasoning with respect to the reference answer.
Given a prompt $x$, a model-generated reasoning segment $y_{\mathrm{think}}$,
and a reference answer $y^\ast$, we compute the token-level negative
log-likelihood under teacher forcing:
\[
\mathrm{NLL}(y^\ast \mid x, y_{\mathrm{think}})
=
-\frac{1}{|y^\ast|}
\log p_\theta\!\left(y^\ast \mid x, y_{\mathrm{think}}\right).
\]
where $p_\theta(y^\ast \mid x, y_{\mathrm{think}})$ denotes the
teacher-forced likelihood of the reference answer. Lower NLL indicates higher conditional likelihood.
To ensure robustness across tasks, we apply rank-based normalization
within each GRPO group:
\[
r_{\mathrm{ppl}} = \mathrm{Norm}\!\left(-\mathrm{NLL}\right),
\]
where higher reward corresponds to lower perplexity.
Although SS-GEN is categorized as a creative writing benchmark, each prompt in the dataset is paired with a reference continuation.
This allows the computation of conditional perplexity with respect to the reference text.

\paragraph{Entropy Reward ($r_{\mathrm{ent}}$)}
To encourage exploration and mitigate mode collapse, we define an
entropy-based reward.
For a generated response $y$, the average token-level entropy is:
\[
H(y) = \frac{1}{|y|}
\sum_{t=1}^{|y|}
\mathbb{E}_{p_\theta(\cdot \mid y_{<t})}
\left[ -\log p_\theta(\cdot \mid y_{<t}) \right].
\]
Entropy values are rank-normalized within each GRPO group:
\[
r_{\mathrm{ent}} = \mathrm{Norm}(H),
\]
assigning higher rewards to responses with greater predictive entropy.

\paragraph{Length Penalty ($r_{\mathrm{len}}$)}
To discourage excessively long responses, we define a length-based reward.
Let $|y|$ denote the response length, $\ell_{\min}$ a soft length
threshold, and $\ell_{\max}$ the maximum allowed length.
The length reward is defined as:
\[
r_{\mathrm{len}}(y) =
\begin{cases}
1, & |y| \le \ell_{\min}, \\
1 - \dfrac{|y| - \ell_{\min}}{\ell_{\max} - \ell_{\min}}, &
\ell_{\min} < |y| \le \ell_{\max}, \\
0, & |y| > \ell_{\max}.
\end{cases}
\]

\paragraph{Semantic Preference Reward ($r_{\mathrm{pref}}$)}
We incorporate a semantic preference reward produced by a pretrained
general-purpose reward model.
Given a response $y$, the reward model outputs a scalar score
$s_{\mathrm{pref}}(y)$ reflecting high-level qualities such as
helpfulness and coherence.
We normalize this score to obtain:
\[
r_{\mathrm{pref}} = \mathrm{Norm}\!\left(s_{\mathrm{pref}}(y)\right).
\]
This reward serves as a supplementary signal and is never used as the
dominant optimization objective.

\section{Training Details}
\label{sec:appendix_training}

This appendix provides implementation details for training the backbone
LLMs under the GRPO framework.
Unless otherwise specified, all GRPO-based methods share the same
training configuration within each dataset to ensure fair comparison.

\subsection{Backbone Models}
\label{sec:appendix_backbone}

We run experiments on two backbone LLMs: Qwen3-8B and
Llama-3.1-8B-Instruct.
All models are trained with full-parameter fine-tuning.
Unless otherwise stated, training is performed in \texttt{bf16} precision.

\subsection{Datasets and Preprocessing}
\label{sec:appendix_data}

To ensure fair comparison and minimize data contamination, we largely
follow the dataset construction and preprocessing pipeline introduced
in prior open-domain GRPO work.
In particular, we adopt the same dataset selection, filtering criteria,
and train--validation--test splits as used by NOVER.

Across all datasets, we prioritize recent benchmarks that require
multi-step reasoning and span diverse domains, including general
reasoning, creative writing, social intelligence, and multilingual
generation.
All datasets are cleaned and filtered to remove low-quality or trivial
samples following the protocols described in NOVER.

\paragraph{Dataset-specific Preprocessing}
For Natural Reasoning, GeneralThoughts, and WebInstruct, we exclude
samples without valid reference answers and remove trivial cases (e.g.,
single-word or purely numerical answers).
For datasets originally formatted as multiple-choice questions, we
convert them into free-form question answering, ensuring that reference
answers include both the correct option and its content.
EmoBench and ToMBench follow their standard task-specific formatting.

\paragraph{Prompt Template Adjustment for Verbose Backbones}
To ensure that performance gains in RL are not confounded by the inherent verbosity of the base model, we observed in preliminary runs that Qwen3-8B produces overly verbose and repetitive \texttt{<think>} segments on several datasets, leading to frequent truncation and format violations.
To mitigate this failure mode, we apply a lightweight instruction-level
constraint for Qwen3-8B on datasets with typically short targets
(e.g., ToMBench and EmoBench), encouraging concise reasoning (e.g., fewer
than five sentences) while keeping the same structured output format.
We do \emph{not} apply this constraint to NR and SS-GEN, where longer
reasoning is often necessary.
Importantly, for each dataset/backbone setting, \textbf{the exact same prompt
template is used for all compared methods (MAESTRO and baselines)}. Fig~\ref{fig:prompt-example} shows an illustrative example of the
instruction-level adjustment applied to Qwen3-8B on datasets with
typically short responses.

\paragraph{Open-domain Nature of the Benchmarks}
The evaluated benchmarks predominantly target open-domain generation, where responses may vary widely in form and content and often admit multiple valid realizations.
While some datasets include questions with reference answers or structured formats (e.g., multiple-choice), a substantial portion of the evaluation involves open-ended reasoning, creative writing, or socially grounded responses, for which reliable rule-based verification is difficult to define consistently.
As a result, we adopt an LLM-as-a-Judge evaluation protocol, following prior work on open-domain RL alignment, to enable uniform and scalable preference-based comparisons across diverse task settings. This evaluation protocol is consistent with prior work on GRPO- and RL-based alignment in open-domain settings, where LLM-based judges are commonly adopted when deterministic verification is unavailable \citep{liu2025nover, guo2025deepseek}.

\begin{figure}[t]
\centering
\begin{tcolorbox}[colback=gray!5, title={Illustrative Prompt Adjustment}]

\textbf{Original:}

Answer the question and return in the following format:

\texttt{<think>} \ldots \texttt{</think>}

\texttt{<answer>} \ldots \texttt{</answer>}

\vspace{0.5em}
\textbf{Modified (used for Qwen3-8B on selected datasets):}

Keep the reasoning extremely brief with fewer than 5 sentences.
Do not include unnecessary details.

Return the answer in the following format:

\texttt{<think>} [BRIEF reasoning] \texttt{</think>}

\texttt{<answer>} \ldots \texttt{</answer>}
\end{tcolorbox}
\caption{Illustrative prompt template modification applied to Qwen3-8B.}
\label{fig:prompt-example}
\end{figure}

\paragraph{Data Splits}
For datasets with existing train--test splits, we retain the original
splits.
For large-scale datasets without predefined splits, we follow prior work
and sample balanced subsets for training, validation, and testing to
ensure distributional consistency.
Dataset statistics after filtering and splitting closely match those
reported in NOVER.

\subsection{GRPO Training Configuration}
\label{sec:appendix_grpo}

We train all GRPO-based methods using the TRL implementation of GRPO.
For each prompt $x$, we sample a group of $G$ candidate responses
$\{y_j\}_{j=1}^{G}$ from the current policy $\pi_\theta$.
Group-wise rewards are used to compute relative advantages for stable
policy optimization.

\paragraph{GRPO Implementation}
Following recent practice, we adopt the default DR-GRPO variant provided
by TRL, which incorporates several stability-oriented improvements over
the original GRPO formulation.
Importantly, this implementation choice is applied consistently to all
GRPO-based baselines and to MAESTRO.
Therefore, performance differences among GRPO-based methods reflect the
effect of reward design and weighting strategies, rather than differences
in the underlying optimization algorithm.

\paragraph{Hyperparameters}
Table~\ref{tab:grpo-hparams} summarizes the main training hyperparameters.
Unless otherwise noted, these settings are kept identical across
baselines on the same dataset and backbone.

We adjust a small number of system-level settings by dataset/backbone to
match response-length characteristics.
For Qwen3-8B, we set the maximum completion length to 2048 for NR and
SS-GEN, to 512 for ToMBench and EmoBench, and to 1024 otherwise; for
Llama-3.1-8B-Instruct we use 1024 across all datasets.
We also set the reference synchronization interval to 6 steps by
default, and to 12 steps for SS-GEN to allow slightly more exploration
in long-form generation.
All methods on the same dataset/backbone share the same settings.

\begin{table}[t]
\centering
\small
\setlength{\tabcolsep}{6pt}
\begin{tabular}{l l}
\toprule
\textbf{Parameter} & \textbf{Value} \\
\midrule
Optimizer & AdamW \\
Learning rate & 1.0e-5 \\
Weight decay & TRL default \\
Warmup ratio / steps & TRL default \\
Scheduler & TRL default \\
Max training epochs & 2 \\
Prompt batch size & 15 \\
Gradient accumulation steps & 1 \\
Number of generations & 5 \\
Max prompt length & TRL default \\
Max generation length & 512-2048 \\
Sampling temperature & 0.8 \\
Top-$p$ / Top-$k$ & TRL default \\
\midrule
KL coefficient $\beta$ & 0.1 \\
Reference sync interval & 6-12 \\
Reference mixup alpha & 0.6 \\
\midrule
Loss type & DR.GRPO \\
Top entropy quantile &  0.2 \\
Importance sampling & Sequence-level \\
Mask truncated completions & True \\
\midrule
Precision & bf16 \\
LoRA & Disabled \\
Training GPUs & 6 GPUs \\
vLLM server GPUs & 2 GPUs \\
\bottomrule
\end{tabular}
\caption{
GRPO training hyperparameters used in our experiments.
All unspecified hyperparameters follow the default settings of the TRL framework.
We adopt TRL's DR-GRPO implementation consistently for all GRPO-based baselines and MAESTRO.
}
\label{tab:grpo-hparams}
\end{table}

\subsection{Reward Aggregation and Normalization}
\label{sec:appendix_reward_norm}

All reward components are computed at the response level.
For each prompt, we obtain a raw reward vector $\mathbf{r}(x,y)\in\mathbb{R}^K$
for each sampled response $y$.
The final scalar reward used by GRPO is denoted as $R(x,y)$.

\paragraph{Group-wise Normalization}
To stabilize optimization across heterogeneous tasks and reward scales,
we apply group-wise normalization within each GRPO group.
In particular, rank-based normalization is applied to selected reward
signals (e.g., perplexity- and entropy-based rewards), mapping values to
$[0,1]$ within each group.
The resulting scalar rewards are then used to compute relative
advantages.

\subsection{GRPO Advantage Computation}
\label{sec:appendix_adv}

Given a group of $G$ sampled responses for the same prompt, we compute
the (scalar) reward for each response and derive a group-relative
advantage by subtracting a group baseline.
We follow the standard GRPO protocol implemented in TRL.
Unless otherwise specified, advantages are computed and normalized
within each group before policy updates.

\subsection{Distributed Training and Implementation}
\label{sec:appendix_dist}

We conduct training on A100 GPUs using distributed data-parallel
training.
We use the \texttt{accelerate} library for multi-GPU synchronization.
Model rollouts dominate the overall training time; therefore, we keep
generation settings fixed across baselines to ensure fair wall-clock
comparison.

\section{Evaluation Protocol}
\label{sec:appendix_eval}

Due to the open-ended nature of the evaluated tasks, constructing
deterministic or rule-based evaluation functions is infeasible.
Therefore, we adopt an LLM-as-a-Judge protocol to assess model outputs
across all datasets.

\subsection{Judge Models}

We use different judge models depending on task characteristics.
For creative writing (SS-GEN) and multilingual generation (OPUS-Books), which
require strong generative and multilingual capabilities, we use
Gemini-2.5-Flash as the judge.
For all other datasets, including general reasoning and social
intelligence benchmarks, we use Qwen3-235B-A32B.
These choices follow standard practice and are motivated by the
strengths of the respective judge models in the target domains.

\subsection{Evaluation Consistency}

For each dataset, we strictly ensure that all compared methods and
backbone models are evaluated using \emph{identical} judge models,
prompts, sampling settings, and comparison protocols.
Thus, performance differences reflect differences in model behavior
rather than evaluation artifacts.

\subsection{Voting Protocol}

For each prompt, we sample multiple responses from each model and
conduct pairwise comparisons using the LLM judge.
Each comparison is repeated three times with independent sampling from
the judge model.
The final preference outcome is determined by majority vote.
This procedure reduces variance and improves the robustness of the
evaluation.

\paragraph{Task-specific Evaluation Prompts}
We employ task-aware evaluation prompts to better reflect the intended
objectives of different benchmarks.
For creative writing (SS-GEN), we use a dedicated creative-writing
evaluation prompt as shown in Fig~\ref{fig:eval-prompt-writing} that emphasizes creativity, coherence, and overall
writing quality.
For general reasoning tasks, we primarily adopt a strict correctness
verification prompt as shown in Fig~\ref{fig:eval-prompt-strict} that evaluates model outputs against reference
answers.
Notably, for Natural Reasoning (NR), which features more open-ended
reasoning styles and multiple valid solution paths, we use a more
lenient correctness prompt as in Fig~\ref{fig:eval-prompt-lenient} that allows for semantically equivalent
answers and partial variations in reasoning.

\paragraph{Robustness to Judge Model Choice}
While we use Gemini-2.5-Flash as the judge for SS-GEN and OPUS-Books due to
their higher demands on generative and multilingual capabilities, we
observe that for all other datasets, evaluation results are largely
insensitive to the choice of judge model.
Specifically, using different large-scale judges yields highly
consistent preference rankings across methods, suggesting that our
reported results are robust to the particular choice of evaluation
model outside of creative writing and multilingual tasks.

\begin{figure}[t]
\centering
\begin{tcolorbox}[colback=gray!5, title={Evaluation Prompt (Creative Writing)}]

You are a writing evaluation model.\\
Your task is to evaluate whether the given answer is a high-quality and
complete response that fully satisfies the writing prompt.\\[0.5em]
Writing Prompt:\\
\texttt{\{q\}}\\[0.5em]
Reference Answer (if provided, do NOT evaluate this, just use for context):\\
\texttt{\{ref\}}\\[0.5em]
Model Answer to Evaluate:\\
\texttt{\{ans\}}\\[0.5em]
Answer the question and return in the following format:\\[0.5em]
\texttt{...}\\
So the result is \{True/False\}
\end{tcolorbox}
\caption{Illustrative evaluation prompt for creative writing tasks.}
\label{fig:eval-prompt-writing}
\end{figure}

\begin{figure}[t]
\centering
\begin{tcolorbox}[colback=gray!5, title={Evaluation Prompt (General Reasoning, Strict)}]
You are an answer verification model.\\
Your task is to determine whether the model's answer is
CORRECT with respect to the gold answer.\\[0.6em]
Question:\\
\texttt{\{q\}}\\[0.6em]
Gold Answer (this is the only correct answer you should use for judgment):\\
\texttt{\{ref\}}\\[0.6em]
Model Answer to Evaluate:\\
\texttt{\{ans\}}\\[0.6em]
Answer the question and return in the following format:\\[0.4em]
\texttt{...}\\
So the result is \{True/False\}
\end{tcolorbox}
\caption{Strict illustrative evaluation prompt for general reasoning tasks.}
\label{fig:eval-prompt-strict}
\end{figure}

\begin{figure}[t]
\centering
\begin{tcolorbox}[colback=gray!5, title={Evaluation Prompt (General Reasoning, Lenient)}]
Please evaluate whether the given answer is correct for the question.\\[0.6em]
Question:\\
\texttt{\{q\}}\\[0.6em]
Reference Answer (not to be evaluated):\\
\texttt{\{ref\}}\\[0.6em]
Given Answer to be Evaluated:\\
\texttt{\{ans\}}\\[0.6em]
Answer the question and return in the following format:\\[0.4em]
\texttt{...}\\
So the result is \{True/False\}
\end{tcolorbox}
\caption{Illustrative evaluation prompt for lenient correctness assessment.}
\label{fig:eval-prompt-lenient}
\end{figure}

\section{Conductor Implementation Details}
\label{sec:appendix_conductor}

This appendix provides implementation details of the Conductor module,
including its architecture, training objective, update schedule, and
hyperparameters.
The Conductor is implemented as a lightweight auxiliary network and
introduces negligible overhead compared to the backbone LLM.

\subsection{Architecture and Forward Computation}

The Conductor is implemented as a lightweight linear prediction head
parameterized by $\phi$.
It takes as input a contextual semantic representation
$h(x,y) \in \mathbb{R}^d$, extracted from the final hidden layer of the
backbone LLM after processing the full prompt--response sequence, and
outputs a $K$-dimensional vector of logits:
\[
\alpha = f_\phi(h(x,y)).
\]

These logits are transformed into a continuous reward weight vector
$w(h) \in \Delta^{K-1}$ via a temperature-controlled softmax:
\[
w_k(h) = \frac{\exp(\alpha_k / \tau)}{\sum_{j=1}^{K} \exp(\alpha_j / \tau)}.
\]
To prevent degenerate solutions, we enforce a minimum probability
constraint and renormalize the resulting weights.

At inference time, the Conductor always produces a continuous weight
vector, enabling smooth trade-offs among multiple reward objectives.
This asymmetric design mirrors the formulation in
Section~\ref{sec:Advantage-Driven Meta-Optimization}:
while training relies on discrete sampling for unambiguous credit
assignment, inference employs the full continuous distribution to
achieve stable multi-objective balancing.

\subsection{Discrete Training Objective}
\label{sec:onehot}

Although the Conductor outputs continuous reward weights during
inference, it is trained using a discrete latent action formulation to
enable stable \emph{rollout-level} credit assignment.
For each sampled rollout $y_{i,j}$ corresponding to prompt $x_i$, the
Conductor samples a reward-orchestration action
$a_{i,j} \sim \pi_\phi(\cdot \mid h_{i,j})$, where
$h_{i,j} = h(x_i, y_{i,j})$.
Each action $a_{i,j} \in \{1, \dots, K\}$ corresponds to emphasizing a
specific reward mode during training.

Following the formulation in
Section~\ref{sec:Advantage-Driven Meta-Optimization}, the Conductor is
optimized using a REINFORCE-style meta-objective.
Given a meta-batch of $N$ prompts, each with $G$ rollouts, the
meta-loss is defined as:
\begin{equation}
\mathcal{L}_{\mathrm{meta}}
=
- \frac{1}{N \cdot G}
\sum_{i=1}^{N} \sum_{j=1}^{G}
\hat{A}_{i,j}
\cdot
\log \pi_\phi(a_{i,j} \mid h_{i,j}),
\end{equation}
where $\hat{A}_{i,j}$ denotes the group-relative GRPO advantage of the
$j$-th rollout for prompt $x_i$, computed under the scalarized reward
induced by $(h_{i,j}, a_{i,j})$.
This formulation ensures that the Conductor receives direct learning signals reflecting the effectiveness of its reward-emphasis decisions at the rollout level.

\subsection{Regularization and Stability}

To encourage exploration over reward emphases and prevent premature
collapse, we augment the meta-loss with an entropy regularization term:
\begin{equation}
\mathcal{L}_{\mathrm{ent}} =
- \lambda_{\mathrm{ent}} \,
\mathcal{H}\!\left(\pi_\phi(\cdot \mid h)\right).
\end{equation}
This regularization delays over-confident reward assignments and
stabilizes learning, particularly during early training.
Temperature scaling and minimum-probability constraints further ensure
that all reward components remain explorable across diverse contexts.

\subsection{Hyperparameters}

The Conductor is optimized using a REINFORCE-style policy gradient
objective with asynchronous updates.
Unless otherwise specified, hyperparameters are fixed across datasets
and backbone models.
We update the Conductor periodically every $M$ GRPO steps.
By default, we use $M=3$, and for SS-GEN we adopt a slower schedule
($M=6$) to match the longer-horizon exploration setting where the
reference model is synchronized less frequently (12 vs.\ 6 steps).
Table~\ref{tab:conductor-hparams} summarizes the hyperparameters used in
all experiments.

\begin{table}[t]
\centering
\small
\setlength{\tabcolsep}{6pt}
\begin{tabular}{l l}
\toprule
\textbf{Parameter} & \textbf{Value} \\
\midrule
Input representation & Final-layer embedding \\
Number of reward objectives ($K$) & 5 \\
Optimizer & AdamW \\
Learning rate & 5e-5 \\
Update interval ($M$) & 3-6 GRPO steps \\
Entropy regularization & Enabled \\
Entropy coefficient $\lambda_{\mathrm{ent}}$ & 1e-3 \\
Softmax temperature $\tau$ & Learned (init=1.0) \\
Minimum probability $\epsilon$ & 1e-4 \\
\bottomrule
\end{tabular}
\caption{
Hyperparameters for the Conductor module.
}
\label{tab:conductor-hparams}
\end{table}

\section{Implementation Details of LLM-as-a-Judge Rewards}
\label{sec:appendix_llm_judge_impl}
\begin{figure}[h]
\centering
\begin{tcolorbox}[colback=gray!5, title=Evaluation Prompt (LLM-as-a-Judge)]
You are a strict but fair evaluation model.
You are given a question, a reference answer, and a candidate response
to be evaluated.\\[0.6em]
Question:\\
\texttt{\{-question-\}}\\[0.6em]
Reference Answer:\\
\texttt{\{-references-\}}\\[0.6em]
Candidate Answer to Evaluate:\\
\texttt{\{-completion-\}}\\[0.6em]
Please assign a score between \textbf{0 and 1} to reflect the quality of
the candidate answer with respect to the question:\\
-- \textbf{0} indicates the answer is completely incorrect or irrelevant;\\
-- \textbf{1} indicates the answer is fully correct, clear, and relevant;\\
-- The score may be any decimal value between 0 and 1
(e.g., 0.0, 0.25, 0.73, 1.0).\\[0.6em]
You must output \textbf{only a single decimal number between 0 and 1}.
Do \emph{not} output any additional text, symbols, or explanations.
Simply output the score itself, for example:\\
\texttt{0.73}
\end{tcolorbox}
\caption{Illustrative LLM-as-a-Judge prompt for GRPO reward.}
\label{fig:eval-prompt-judge-score}
\end{figure}
This section provides implementation details for the LLM-as-a-Judge
reward baseline reported in the main paper.
The purpose is to clarify how the judge-based reward is integrated into
the GRPO pipeline, rather than to introduce additional experimental
comparisons.

We implement the LLM-as-a-Judge reward within the same GRPO training
framework used by MAESTRO.
All training hyperparameters, rollout settings, reference model updates,
and optimization schedules are kept identical.
The only difference lies in the reward computation.

Specifically, for each prompt, we first filter model outputs to retain
only responses that satisfy the required output format.
Each valid response is then evaluated by a large language model judge
using the same task-specific evaluation prompts described in
Fig~\ref{fig:eval-prompt-judge-score}.
The judge produces a continuous score in the range $[0,1]$, reflecting
the overall quality or correctness of the response.

Following standard GRPO practice, these judge scores are converted into
scalar rewards by ranking responses within each rollout group and
applying rank-based normalization.
The resulting normalized rewards are then used for advantage estimation
and policy optimization.

This implementation isolates the effect of the reward signal itself,
ensuring a fair comparison without introducing additional differences
in training dynamics or evaluation protocols.

\section{Group-Relative Policy Optimization (GRPO) Details}
\paragraph{GRPO Objective}
Given a prompt $x$, GRPO samples a group of $G$ responses $\{y_j\}_{j=1}^G$ from the policy $\pi_\theta$.
For each response, a scalar reward $R(x, y_j)$ is computed, and group-relative advantages are obtained as:
\[
\hat{A}_j = \frac{R(x,y_j) - \mu_R}{\sigma_R},
\]
where $\mu_R$ and $\sigma_R$ denote the mean and standard deviation of rewards within the group.

GRPO optimizes the policy by maximizing group-relative advantages with KL regularization:
\[
\mathcal{L}_{\mathrm{GRPO}}(\theta)
=
\mathbb{E}_{x,y \sim \pi_\theta}
\left[
\hat{A}(x,y)
\right]
-
\beta \, \mathrm{KL}\!\left(
\pi_\theta \,\|\, \pi_{\mathrm{ref}}
\right).
\]
Here, the KL term penalizes deviation from a reference policy $\pi_{\mathrm{ref}}$,
serving as a trust-region constraint to stabilize policy updates.

\paragraph{Group-Relative Advantages}
In GRPO, rewards are normalized within each rollout group before computing advantages:
\[
\hat{A}_{i,j} = \frac{R_{i,j} - \mu_i}{\sigma_i},
\]
where $\mu_i$ and $\sigma_i$ denote the mean and standard deviation of rewards for prompt $x_i$.
This group-wise normalization removes prompt-dependent reward scale and bias,
so policy updates depend only on the relative quality of responses generated for the same prompt.
As a result, GRPO achieves stable optimization even when reward magnitudes vary substantially across tasks or inputs

\section{Use of AI Assistance}
\label{sec:ai_assistance}

We used AI-based tools to assist with language editing and clarity during the writing of this paper.
The AI assistance was limited to improving phrasing, grammar, and overall readability, and did not contribute to the scientific content of the work.
All research ideas, methodological design, experiments, analyses, and conclusions were developed solely by the authors.

\end{document}